\documentclass[10pt,twocolumn,letterpaper]{article}

\usepackage{iccv}
\usepackage{times}
\usepackage{epsfig}
\usepackage{graphicx}
\usepackage{amsmath}
\usepackage{amssymb}
\usepackage{multirow}
\usepackage{booktabs}
\usepackage[accsupp]{axessibility}  

\newcommand{\norm}[1]{\left\lVert#1\right\rVert}
\newcommand{\abs}[1]{\left\lvert#1\right\rvert}

\usepackage[pagebackref=true,breaklinks=true,letterpaper=true,colorlinks,bookmarks=false]{hyperref}

\usepackage[capitalize]{cleveref}

\iccvfinalcopy 

\def\iccvPaperID{10960} 

\begin{document}

\crefname{section}{Sec.}{Secs.}
\Crefname{section}{Section}{Sections}
\Crefname{table}{Table}{Tables}
\crefname{table}{Tab.}{Tabs.}

\title{4D Panoptic Segmentation as Invariant and Equivariant Field Prediction}

\author{
Minghan Zhu$^{1,}$\thanks{Work done at Qualcomm AI Research during an internship.}
\quad\quad
Shizhong Han$^2$
\quad\quad
Hong Cai$^2$
\quad\quad
Shubhankar Borse$^2$
\and
Maani Ghaffari$^{1,*}$
\quad\quad
Fatih Porikli$^2$\\\\
{$^1$University of Michigan, Ann Arbor
\quad\quad
$^2$Qualcomm AI Research\thanks{Qualcomm AI Research, an initiative of Qualcomm Technologies, Inc.} }
}
\maketitle

\newcommand{\hc}[1]{\textcolor{magenta}{[HC: #1]}}
\newcommand{\mhz}[1]{\textcolor{blue}{[Minghan: #1]}}


\begin{abstract}
In this paper, we develop rotation-equivariant neural networks for 4D panoptic segmentation. 4D panoptic segmentation is a benchmark task for autonomous driving that requires recognizing semantic classes and object instances on the road based on LiDAR scans, as well as assigning temporally consistent IDs to instances across time. We observe that the driving scenario is symmetric to rotations on the ground plane. Therefore, rotation-equivariance could provide better generalization and more robust feature learning. Specifically, we review the object instance clustering strategies and restate the centerness-based approach and the offset-based approach as the prediction of invariant scalar fields and equivariant vector fields. Other sub-tasks are also unified from this perspective, and different invariant and equivariant layers are designed to facilitate their predictions. Through evaluation on the standard 4D panoptic segmentation benchmark of SemanticKITTI, we show that our equivariant models achieve higher accuracy with lower computational costs compared to their non-equivariant counterparts. Moreover, our method sets the new state-of-the-art performance and achieves 1st place on the SemanticKITTI 4D Panoptic Segmentation leaderboard. 

\end{abstract}

\section{Introduction}\label{sec:intro}
Perception with LiDAR point clouds is an important part of building real-world autonomous systems, for example, self-driving cars~\cite{sun2020scalability, qian20223d, ZhongZP21, shi2019pairwise}. As the computer vision community gradually builds more capable neural networks, the tasks also become more complex. 4D panoptic segmentation~\cite{aygun20214d} is an emerging task that combines several previously independent tasks: semantic segmentation, instance segmentation, and object tracking, in a unified framework, given sequential LiDAR scans. As the output provides abundant useful information for understanding the dynamic driving environment, solving this task has significant practical value.  

\begin{figure}
    \centering
    \includegraphics[width=\columnwidth]{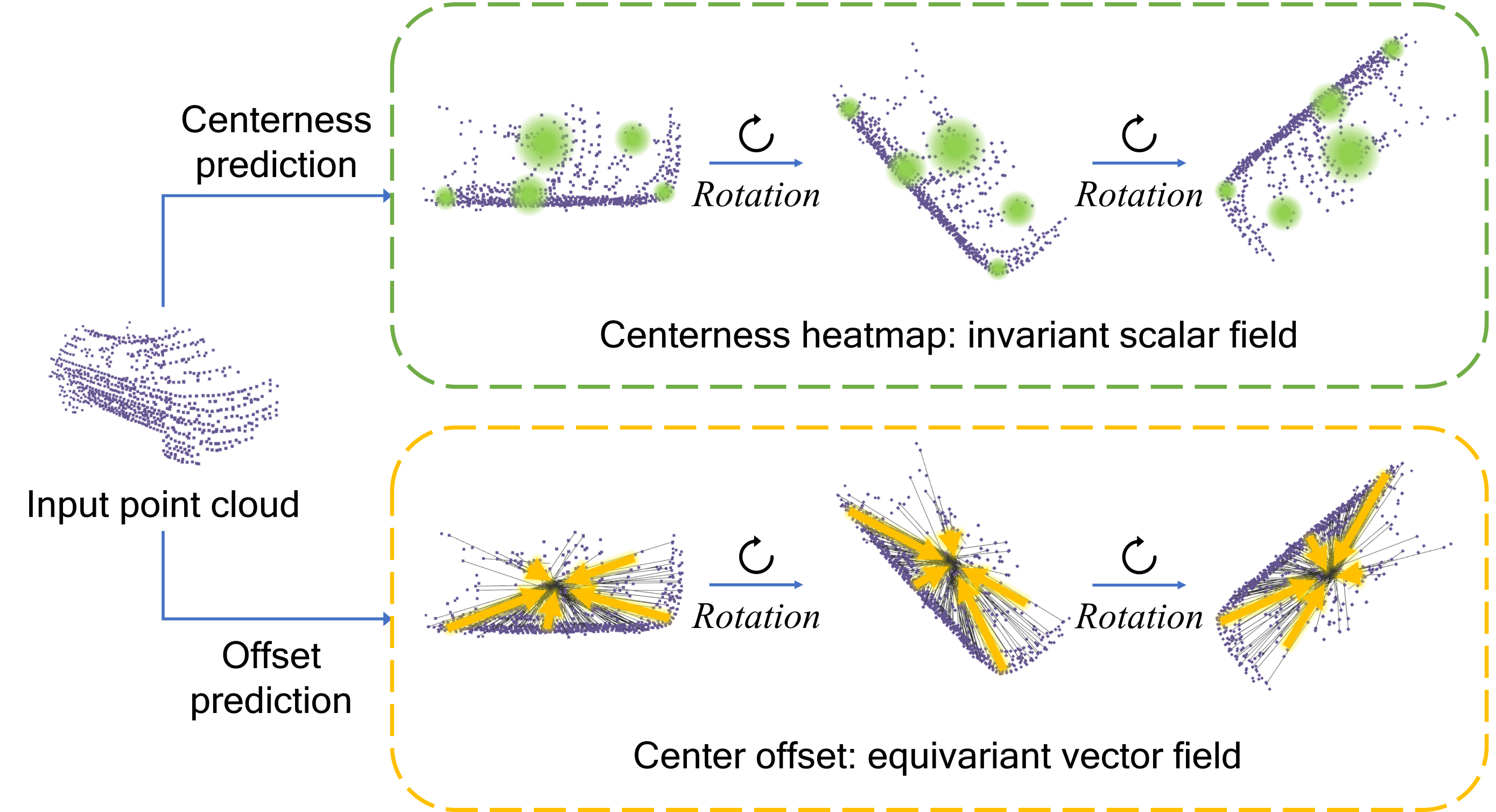}
    \caption{Predicting centerness or offset is a critical step in estimating object centers as part of the overall panoptic segmentation. A centerness heatmap can be viewed as an invariant field, while the offset can be viewed as an equivariant vector field. In our proposed approach, we devise different, corresponding methods to predict invariant and equivariant fields, respectively.}
    \label{fig:ctr_vs_off}
\end{figure}

While there are a few existing methods to solve 4D panoptic segmentation~\cite{aygun20214d, kreuzberg20234d}, they ignore the rich, inherent symmetries present in this task. For example, for the point cloud of an object instance, its center is invariant to rotations, and the offset vector from any point on the object to the center is attached to and thus rotates along with the body frame. See Fig.~\ref{fig:ctr_vs_off} for a visual illustration.

As such, in this paper, we propose to develop equivariant neural networks to solve 4D panoptic segmentation. Equivariant networks~\cite{cohen2016group} are deep learning models that are guaranteed to render outputs that respect the symmetries in data. For the 4D panoptic segmentation task, rotational equivariance can help the model perform consistently and generalize over rotations in the input data. 

While equivariance is a nice property, equivariant models can be complex and incur high computational costs~\cite{zhu20222, yu2022rotationally, wu2022transformation}. As a result, most existing equivariant models are only applied to small-scale problems, such as molecular analysis and single-object perception \cite{fuchs2020se}. 
Recent works have looked into more efficient equivariant networks~\cite{zhu20222} and applications in larger problems \cite{yu2022rotationally}, but significant performance improvement without extra computational cost has not been achieved in large-scale equivariant perception solutions. 

In this work, SO(2)-equivariance is incorporated in the 4D segmentation model. 
We find that the equivariance brings consistent improvements in several performance metrics and that formulating the output as equivariant vector fields helps maximize the benefits of equivariant models, as compared to restricting to invariant scalar fields (see \cref{fig:ctr_vs_off}). 
Furthermore, our equivariant networks can improve the segmentation performance while reducing computational costs at the same time. With our proposed design, we outperform the non-equivariant models and, notably, achieve the \textbf{top-1} ranking position in the SemanticKITTI benchmark.  

Our main contributions in this paper are as follows:
\begin{itemize}
    \item We develop the first rotation-equivariant model for 4D panoptic segmentation, bringing improvements in both performance and efficiency.
    \item We investigate different strategies and designs to construct the equivariant architecture. Specifically, we discover the advantage of formulating prediction targets as equivariant vector fields, as compared to only invariant fields.
    \item Evaluated on the SemanticKITTI benchmark, our equivariant models significantly outperform 
    existing methods, validating the value of equivariance in this large-scale perception task. 

    \item Our code is open-sourced at \url{https://eq-4d-panoptic.github.io/}. 
\end{itemize}

\section{Related work} \label{sec:literature}

\subsection{LiDAR 3D and 4D Panoptic segmentation}\label{sec:lit_seg}
\paragraph{3D Panoptic segmentation}
The task of panoptic segmentation is first proposed in the image domain \cite{kirillov2019panoptic} and later extended to LiDAR point clouds with the release of a large-scale outdoor LiDAR point cloud dataset with panoptic labels, SemanticKITTI \cite{behley2021benchmark}. Similar to the semantic segmentation~\cite{borse2021inverseform, borse2021hs3, hu2022learning, borse2023dejavu, borse2023x} and panoptic segmentation techniques in the image domain \cite{mohan2021efficientps, yang2019deeperlab, borse2022panoptic}, their 3D counterparts can be classified into proposal-based and proposal-free methods. Proposal-based methods \cite{behley2021benchmark, sirohi2021efficientlps} require a detection module to locate the objects first and then predict the instance mask for each bounding box and conduct semantic segmentation on the background pixels. This strategy needs to deal with potential conflicts among the segmentations. 
On the other hand, proposal-free methods conduct semantic segmentation first and then cluster the points belonging to different instances. The clustering strategy impacts overall efficiency and performance. Offset prediction and centerness prediction are two major approaches for this. \textit{Offset prediction} \cite{hong2021lidar, xu2022sparse} means that each point predicts the offset vector to the instance center, and the clustering is conducted on the predicted centers. \textit{Centerness prediction} \cite{aygun20214d} is to regress a heatmap of the closeness to the instance center at any spatial location. Then the local maximums on the heatmap are used to cluster the points nearby. These two strategies can also be combined \cite{zhou2021panoptic, li2022panoptic}. Other clustering methods also exist. For example, \cite{gasperini2021panoster} proposes an end-to-end clustering layer. \cite{razani2021gp} uses a graph network to cluster over-segmented points into instances. 

\paragraph{4D Panoptic segmentation}
The 4D task is to provide temporally consistent object IDs on top of 3D panoptic segmentation. MOPT \cite{hurtado2020mopt} is an early attempt to provide tracking ID for panoptic outputs. 4D-PLS \cite{aygun20214d} proposes evaluation metrics and a strong baseline method for this task. It accumulates point clouds in sequential timestamps to a common frame, applies segmentation on the aggregated point clouds, and clusters the instances using the centerness prediction. 4D-DS-Net \cite{hong2022lidar} and 4D-StOP \cite{kreuzberg20234d} follow this pipeline but use offset prediction to cluster the instances. CA-Net \cite{marcuzzi2022contrastive} uses an off-the-shelf 3D panoptic segmentation network and learns the temporal instances association through contrastive learning. 

\subsection{Equivariant Learning}
\paragraph{Equivariant neural networks}
The equivariance to translations enables CNNs to generalize over translations of image content with much fewer parameters compared with fully connected networks. Equivariant networks extend the symmetries to rotations, reflections, permutations, etc. G-CNN \cite{cohen2016group} enables equivariance to 90-degree rotations of images. Steerable CNNs extend the symmetry to continuous rotations \cite{weiler2018learning}. The input is also extended from 2D images to spherical \cite{cohen2018spherical, esteves2018learning} and 3D data \cite{weiler20183d}. To deal with infinite groups (e.g., continuous rotations), generalized Fourier transforms and irreducible representations are adopted to formulate convolutions in the frequency domain \cite{worrall2017harmonic, thomas2018tensor}. Equivariant graph networks \cite{satorras2021n} and transformers \cite{fuchs2020se} are also proposed as non-convolutional equivariant layers. Equivariant models have applications in various areas such as physics, chemistry, and bioimaging \cite{thomas2018tensor, fuchs2020se}, where symmetries play an important role. They also attract research interest in robotic applications. 
For example, equivariant networks with SO(3)- and SE(3)-equivariance are developed to process 3D data such as point clouds \cite{deng2021vector, chen2021equivariant}, meshes \cite{de2020gauge}, and voxels \cite{weiler20183d}. However, due to the added complexity, most equivariant networks for 3D perception are restricted to relatively simple tasks with small-scale inputs, such as object-wise classification, registration, part segmentation, and reconstruction \cite{sajnani2022condor, zhu2022correspondence, chatzipantazis2022se}. In the following, we will review recent progress in extending equivariance to large-scale outdoor 3D perception tasks. 

\paragraph{Equivariant networks in LiDAR perception}
LiDAR perception demands two main considerations in equivariant models. Firstly, outdoor scene perception typically needs a sophisticated network design, and incorporating equivariant layers that are compatible with conventional network layers can build on existing successful designs. Secondly, due to the large sizes of LiDAR point clouds, it's essential to create expressive equivariant models that fit within the memory constraints of standard GPUs.

Existing work mainly follows two strategies. With the first strategy, inputs are projected to rotation invariant features using local reference frames \cite{li2023improving, xie2023general}. In this way, the changes are mainly at the first layer of the network, causing limited memory overhead. The main drawback is that the invariant feature could cause information loss and limit performance. The second strategy adopts group convolution, by augmenting the domain of feature maps to include rotations \cite{yu2022rotationally, wu2022transformation}. While achieving improved performance with the help of equivariance, they consume twice \cite{wu2022transformation} or four times \cite{yu2022rotationally} of memory as their non-equivariant counterparts due to the augmented dimension of the feature maps and convolutions. With the two strategies, equivariant networks have been applied in 3D object detection \cite{yu2022rotationally, wu2022transformation, xie2023general} and semantic segmentation \cite{li2023improving}.

\section{Equivariance for 4D Panoptic Segmentation}
In this section, we provide preliminaries on equivariance and formulate the dense prediction task from the perspective of equivariant learning. We discuss how to learn equivariant features and fit equivariant prediction target fields, which leads to the design choices in our proposed architecture.

\subsection{Preliminaries on Equivariance}
\paragraph{Feature maps as fields}
A feature map $f_0$ is a \textit{field}, assigning a value to each point in some space. In the context of point cloud perception, we have $f_0: \mathbb{R}^3\rightarrow V$, where $V$ is some vector space. 
This map can represent the geometry of the point cloud, i.e., $f_0(x)=1$ for every point $x$ in the point cloud and $f_0(x)=0$ otherwise. It can also represent point properties or arbitrary learned point features. For example, the feature map $f_0: x \mapsto SC(x)$, where $SC(x)$ is the semantic class label of the point $x$, represents the semantic segmentation of a point cloud. We denote the space of all such feature maps as $\mathcal{F}_0$. 

\paragraph{Invariant and equivariant fields}
For the \textit{group} of transformations $G$, we use the rotation group $\mathrm{SO}(3)$ for the formulation, for which $\mathrm{SO}(2)$ is also valid. 

A rotation $R\in \mathrm{SO}(3)$ can be applied to a point and to a feature map. A \textit{rotated feature map} $[Rf_0]$ is simply the feature map of the rotated point cloud. A point at $x$ goes to $Rx$ after the rotation, thus $f_0(x)$ and $[Rf_0](Rx)$ are the features of the same point before and after rotating the point cloud. The relation between them depends on the property of $V$. For instance, if $V=SC$, then we know 
\begin{equation}\label{eq:inv}
    [Rf_0](Rx)=f_0(x),\quad \forall R\in\mathrm{SO}(3),
\end{equation}
i.e., the rotation does not change the semantic class.

As another example, if $V$ represents the surface normal vector, then we have 
\begin{equation}\label{eq:equiv}
    [Rf_0](Rx)=Rf_0(x),\quad \forall R\in\mathrm{SO}(3),
\end{equation}
where on the right-hand side, $R$ is applied to $f_0(x)\in\mathbb{R}^3$, meaning that the normal vector of a given point rotates along with the point cloud.

In both cases, the feature map of a rotated point cloud, $[Rf_0]$, can be generated from $f_0$, the feature map of the original point cloud. This property is called \textit{equivariance}. In the case of \cref{eq:inv}, we call $f_0$ an \textit{invariant scalar field}. In the case of \cref{eq:equiv}, we call $f_0$ an \textit{equivariant vector field}. 

\paragraph{Learning equivariant features}
A dense prediction task on a point cloud is to reproduce a target field (e.g., $x\mapsto SC(x)$) using a feature map realized by a neural network. Naturally, it would be helpful to equip the network with the same invariant and/or equivariant properties as the target field. However, a general feature map learned by a network is typically neither invariant nor equivariant to rotations. To fix this, we can augment the space of feature maps to $\mathcal{F}=\{f: \mathbb{R}^3\times \mathrm{SO}(3)\rightarrow V\}$, defined by
\begin{equation}\label{eq:def}
    f(x, R) := [R^{-1}f_0](R^{-1}x)
\end{equation}
for some $f_0\in \mathcal{F}_0$, i.e., the augmented feature map at rotation $R$ equals the original feature map rotated by $R^{-1}$. In this way, we have, $\forall R, R'\in \mathrm{SO}(3)$, 
\begin{equation}\label{eq:eq}
    [Rf](Rx, R')\!=\![{R'}^{-1}\!R f_0]({R'}^{-1\!}Rx)\!=\!f(x, R^{-1}\!R'),
\end{equation}
which means that the augmented feature map of a rotated point cloud, $[Rf]$, can be generated by the augmented feature map of the original point cloud $f$, indicating that $f$ is equivariant. Equivariant feature maps satisfying \cref{eq:def} can be constructed using group convolutions~\cite{cohen2016group, chen2021equivariant, zhu20222}.

\subsection{Fitting Equivariant Targets} \label{sec:fitinvequiv}
Now we can use the learned equivariant feature map $f$ to fit the target field $f_{gt}\in \mathcal{F}_0$, which is invariant or equivariant. 
Suppose that we can fit $f_{gt}$ using $f$, then $[Rf]$, the feature map of a rotated point cloud, should automatically fit $[Rf_{gt}]$ which is the target field of the rotated point cloud. In other words, equivariance enables the \textit{generalization} over rotations. Next, we introduce two strategies to achieve this.

\paragraph{Rotational coordinate selection}
Assuming that we can fit the target field of a point cloud without rotation: $f(x, I) = f_{gt}(x)$, where $I$ is identity rotation. We want to show that the fitting generalizes to the rotated point clouds. 

If $f_{gt}$ is an invariant scalar field, from \cref{eq:eq}, we have
\begin{equation}
    [Rf](Rx, R) = f(x, I) = f_{gt}(x) = [Rf_{gt}](Rx),
\end{equation}
meaning that the feature map $[Rf]$ at the rotational coordinate $R$, $[Rf](\cdot, R)$, fits the target $[Rf_{gt}]$ for the rotated point cloud. 

If $f_{gt}$ is an equivariant vector field, we have
\begin{equation}
    [Rf](Rx, R)\! =\! f(x, I)\! =\! f_{gt}(x)\! =\! R^{-1}[Rf_{gt}](Rx),
\end{equation}
which means that we need to apply a rotational matrix multiplication on features in $[Rf](\cdot, R)$ to fit $[Rf_{gt}]$, i.e., $[Rf_{gt}](Rx) = R[Rf](Rx, R), \forall x\in \mathbb{R}^3$.

The analysis above shows that the fitting of $f\in\mathcal{F}$ to $f_{gt}\in\mathcal{F}_0$ generalizes over rotations, if the rotational coordinate (i.e., the second argument in $f$) is the same as the actual rotation of the point cloud. 
However, the actual rotation $R$ is usually \textit{unknown} during inference; thus needs to be learned. In practice, this is formulated as a \textit{rotation classification} task to select the best rotational channel in a feature map $f$. We will discuss how we instantiate this in the architecture in \cref{sec:head}. 

\paragraph{Invariant pooling layer}
When the target field $f_{gt}$ is invariant, there is another fitting strategy. For equivariant feature map $f$ that satisfies \cref{eq:def}, we can build a rotation-invariant layer by marginalizing over the second argument of $f$, i.e.,
\begin{equation}
    f_{inv}(x) = \prod_R f(x, R),
\end{equation}
where $\prod$ denotes any summarizing operator symmetric to all its arguments, e.g., sum, average, and max.

If we can fit $f_{inv}(x)=f_{gt}(x)$, then based on \cref{eq:eq}, we have
\begin{equation}
\begin{split}
    [Rf_{inv}](Rx) &= \prod_{R'} [Rf](Rx, R') = \prod_{R'} f(x, R^{-1}R') \\
    &= f_{inv}(x) = f_{gt}(x) = [Rf_{gt}](Rx),
\end{split}
\end{equation}
which indicates that generalization over rotations holds. In \cref{sec:head}, we discuss different options of implementing the invariant layer in the network. 

\subsection{Equivariant Instance Segmentation}
For instance segmentation, the prediction targets can be modeled as invariant or equivariant fields. As discussed in \cref{sec:lit_seg}, centerness regression and offset regression are two predominately used approaches to estimate the object centers. 
Specifically, we can see that \textit{centerness is an invariant scalar field} and \textit{offset is an equivariant vector field}. Accordingly, we propose different prediction layers in \cref{sec:head}, resulting in different performances shown in \cref{sec:exp_res}.

\section{Proposed Network Architecture}\label{sec:method}

\begin{figure*}
    \centering
    \includegraphics[width=\textwidth]{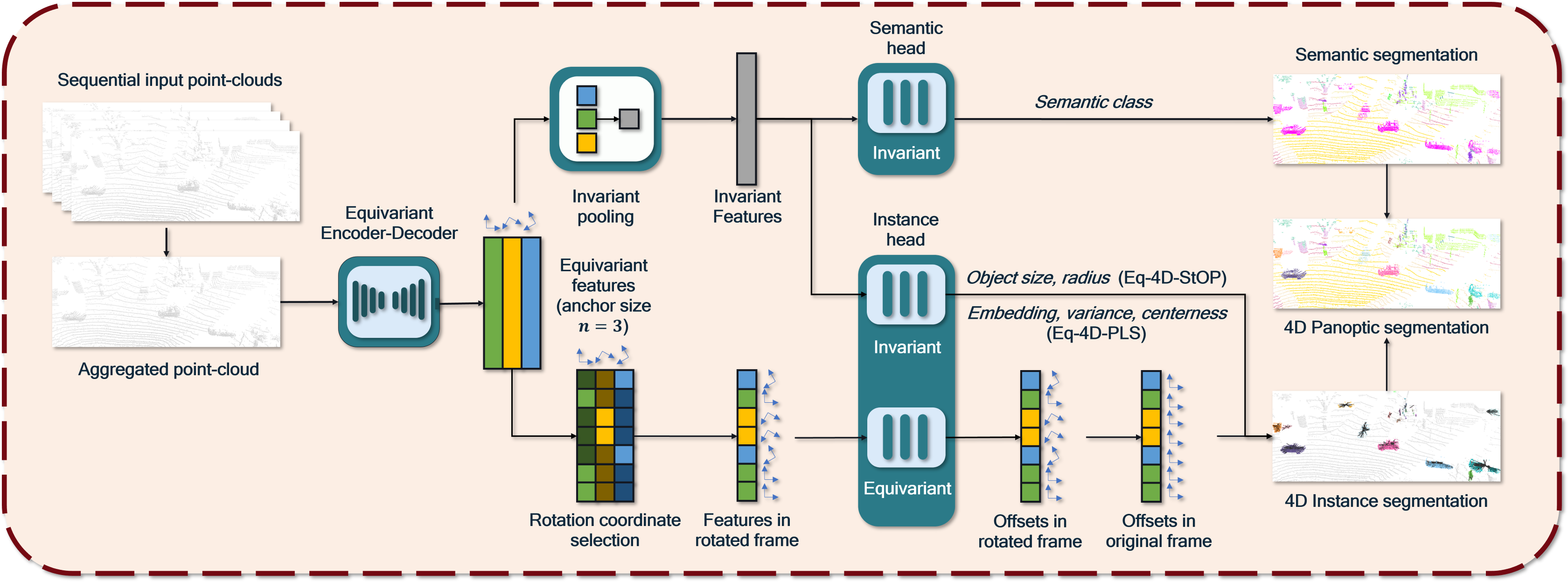}
    \caption{Overview of the network structure. All prediction targets can be classified as equivariant (e.g., offsets to object centers) and invariant (e.g., semantic classes). The gray color represents rotation-invariant features, which is used for invariant predictions. The green, yellow, and blue colors represent features at different rotational coordinates depicted by the small frames. Equivariant predictions use features from the rotational coordinate selection. Whether a rotation is selected is illustrated using the light and dark shades of colors. }
    \label{fig:overview}
\end{figure*}

\subsection{Network Overview}
\paragraph{Discretized SO(2)-equivariance} 

The model's group equivariance should match the data's actual transformations. An overly large group may lead to high computational costs with minimal performance improvement. For the outdoor driving scenario, the $\mathrm{SO}(2)$ group is chosen to represent planar rotations around the gravity axis.

We discretize $\mathrm{SO}(2)$ into a finite group and create a group CNN that is equivariant to the discretized rotations rather than the continuous ones. It allows us to leverage simpler structures akin to conventional deep learning models, allowing integration with existing state-of-the-art networks. Despite discretization, the network is expected to interpolate the equivariance gaps through training with rotational data augmentation.

Specifically, $\mathrm{SO}(2)$ is discretized into cyclic groups $C_n$, where $n$ indicates the number of discretized rotations, such as $C_3$ for 120-degree rotations. These are referred to as the \textit{rotation anchors}.

\paragraph{Network structure}
We utilize the point-convolution style equivariant network E2PN~\cite{zhu20222} as our backbone, which is an equivariant version of KPConv \cite{thomas2019kpconv}. We describe necessary adaptations to E2PN in \cref{sec:encdec}, which allow us to build equivariant models on top of SOTA 4D panoptic segmentation network 4D-PLS~\cite{aygun20214d} and 4D-StOP~\cite{kreuzberg20234d}, both of which are based on KPConv. We refer to our equivariant models as \textit{Eq-4D-PLS} and \textit{Eq-4D-StOP}, respectively. 

On a high level, both models first stack the point clouds from several sequential time instances within a common reference frame so that the temporal association becomes part of the instance segmentation. Each network consists of an encoder, a decoder, and prediction heads. 
The encoders and decoders are very similar for the two models, while the main differences lie in their prediction heads, which formulate the targets as invariant scalar fields and equivariant vector fields, respectively, as discussed in \cref{sec:head}.  
An overview of our network structure is shown in \cref{fig:overview}. 

\subsection{Equivariant Encoder and Decoder}\label{sec:encdec}
\paragraph{Equivariant encoder}
The encoder of the 4D panoptic segmentation networks can be made equivariant by simply swapping the KPConv \cite{thomas2019kpconv} layers with E2PN \cite{zhu20222} convolution layers. However, E2PN is originally designed for $\mathrm{SO}(3)$ equivariance, and uses quotient representations and efficient feature gathering to improve the efficiency. Using it for $\mathrm{SO}(2)$ equivariance requires two adaptations. 

First, we use the regular representation instead of the quotient representation in~\cite{zhu20222}. This is because $\mathrm{SO}(2)$ is abelian, in which case quotient representations cause loss of information (see \cref{sec:quo} for details). 

Second, to apply the efficient feature gathering in E2PN, the spatial position of the convolution kernel needs to be symmetric to the rotation anchors. It allows the rotation of kernel points to be implemented as a permutation of their indices. This implies that different $n$'s in $C_n$ impose different constraints on the number and distribution of the kernel points. For example, the default KPConv kernel with 15 points is symmetric to 60-degree rotations and thus can be used to realize $C_2$, $C_3$, and $C_6$ equivariance. However, $C_4$ requires a different kernel (for symmetry to 90-degree rotations), for which the 19-point KPConv kernel works. 

\paragraph{Equivariant decoder}
The original E2PN only provides an encoder to predict a single output for an input point cloud. We need to devise an equivariant decoder for the dense 4D panoptic segmentation. Similar to conventional non-equivariant networks, our decoder adopts upsampling layers and 1-by-1 convolution layers. The upsampling layers simply assign the features of the coarser point cloud to the finer point cloud via nearest neighbor interpolation. The 1-by-1 convolution processes the feature at each point and each rotation independently. It is straightforward to prove the equivariance of such a decoder (see \cref{sec:proof}).

\subsection{Equivariant Prediction Heads}\label{sec:head}
Our baseline models 4D-PLS~\cite{aygun20214d} and 4D-StOP~\cite{kreuzberg20234d} have different prediction heads. Their semantic segmentation heads are similar, but they employ different clustering approaches for instance segmentation. Correspondingly, we propose different equivariant prediction designs for them.

\subsubsection{Eq-4D-PLS: Segmentation as Invariant Scalar Field Prediction}\label{sec:seg_inv}
In 4D-PLS~\cite{aygun20214d}, instance segmentation is done by clustering the point embeddings, assuming a Gaussian distribution for the embeddings of each instance. It also predicts a point-wise centerness score, measuring the closeness of a point to its instance center, which is used to initialize the cluster centers. Both the point embeddings and the centerness scores can be viewed as \textit{invariant scalar fields}. Note that while these targets can appear like a vector, they are actually a stack of scalars invariant to rotations. 

As discussed in \cref{sec:fitinvequiv}, there are two strategies to fit invariant targets, i.e., rotation coordinate selection and invariant pooling. For invariant pooling layers, max pooling and average pooling over the rotational dimension are two obvious choices. 
For the rotational coordinate selection strategy, a key challenge is that the ground truth for the rotational coordinate $R$ is unavailable or even undefined (as there is no canonical orientation for a LiDAR scan). In addition, existing dataset, e.g., SemanticKITTI, does not provide object bounding box annotations. Hence the object orientations are also unknown. We use an unsupervised strategy to address this issue. Instead of picking the best rotational dimension, we perform a weighted sum of all rotational dimensions, which is differentiable and allows the model to learn the weight for different rotations. This is equivalent to the group attentive pooling layer in EPN~\cite{chen2021equivariant}. 

In summary, we study three designs, i.e., max pooling, average pooling, and group attentive pooling, for the prediction of the invariant targets, including semantic classes, point embeddings, embedding variances, and centerness scores. 

\subsubsection{Eq-4D-StOP: Segmentation as Equivariant and Invariant Field Prediction}\label{sec:seg_equiv}
4D-StOP~\cite{kreuzberg20234d} uses an offset-based clustering method for instance segmentation. The network predicts an offset vector to the instance center for each input point, creating an \textit{equivariant vector field}. The predicted center locations are clustered into instances, and features from points in the same instance are combined to predict instance properties.

As discussed in \cref{sec:fitinvequiv}, we fit equivariant vector fields through rotational coordinate selection. While we do not have ground-truth rotations, the vector field of offsets naturally defines orientations. Denote an offset vector at point $x$ as $v(x) = x_{ctr}-x\in V=\mathbb{R}^3$, where $x_{ctr}$ is the center of the instance that $x$ belongs to. We can define its rotation in $\mathrm{SO}(2)$ as $\theta(v)=\mathrm{atan2}(v_Y, v_X)$, where $X,Y$ are axes in the horizontal plane and $Z$ is the vertical axis. $\theta$ can be assigned to a discretized rotation coordinate (anchor) $R_{i}^{gt}$ by nearest neighbor.
In this way, each point has a rotation label based on its relative position to the instance center, which is well-defined and equivariant to the point cloud rotations. 

Given the rotation label, we train the network to predict it as a classification task. Given the feature map at each rotational coordinate, a rotation score is predicted: 
\begin{equation}
    s_x(R_i) = \phi(f(x, R_i)),\quad \forall R_i\in \mathrm{SO}(2)',
\end{equation}
where $x\in \mathbb{R}^3$, $\mathrm{SO}(2)'$ is the set of rotation anchors, i.e., $\mathrm{SO}(2)'\cong C_n$, $i=1, ..., n$, and $\phi$ is a scoring function. We concatenate $s_x(R_i)$'s as $S_x=[s_x(R_1), ..., s_x(R_n)]$ and apply a cross-entropy loss function on $S_x$ with label $R_i^{gt}$.  

The semantic segmentation, object size, and radius regression in 4D-StOP are invariant fields. However, note that a key difference from the prediction in Eq-4D-PLS (c.f.~\cref{sec:seg_inv}) is that we now have the rotation labels. As such, the rotational coordinate selection strategy can be applied to predicting the invariant targets. Therefore, for Eq-4D-StOP, we study four options for the invariant target prediction, including max pooling, average pooling, group attentive pooling, and rotational coordinate selection.

\section{Experiments}

\subsection{Experimental setup}
\textbf{Dataset:} We conduct our experiments primarily on the SemanticKITTI dataset \cite{behley2019semantickitti}. The SemanticKITTI dataset establishes a benchmark for LiDAR panoptic segmentation \cite{behley2021benchmark}. It consists of 22 sequences from the KITTI dataset. 10 sequences are used for training, 1 for validation, and 11 for testing. In total, there are 43,552 frames. The dataset includes 28 annotated semantic classes, which are reorganized into 19 classes for the panoptic segmentation task. Among the 19 classes, 8 are classified as \textit{things}, while the remaining 11 are categorized as \textit{stuff}. Each point in the dataset is assigned a semantic label, and for points belonging to things, a temporally consistent instance ID. 

\textbf{Metrics:} The core metric for 4D panoptic segmentation is $LSTQ=\sqrt{S_{cls}\times S_{assoc}}$, which is the geometric mean of the semantic segmentation metric $S_{cls}$ and the instance segmentation and tracking metric $S_{assoc}$. $S_{cls}=\frac{1}{C}\sum_{c=1}^C IoU(c)$ is the segmentation IoU averaged over all semantic classes. The average IoU for points belonging to things and stuff are denoted $IoU^{Th}$ and $IoU^{St}$, respectively. $S_{assoc}$ measures the spatial and temporal accuracy of segmenting object instances. See \cite{aygun20214d} for more details. 

\textbf{Architecture:} For our Eq-4D-PLS and Eq-4D-StOP models, we keep the architectures unchanged from their 4D-PLS~\cite{aygun20214d} and 4D-StOP~\cite{kreuzberg20234d} baselines except for the added rotational coordinate selection and invariant pooling layers necessary for equivariant and invariant field predictions. The input size, the batch size, and the learning rate also follow the baselines, respectively. 

As the efficiency (especially the memory consumption) is a pain point for equivariant learning in large-scale LiDAR perception tasks, we specify the number of channels (network width) $c$ and the rotation anchor size $n$ in our analysis. The size of an equivariant feature map $f$ is $\abs{f}=mcn$ for a point cloud with $m$ points. Non-equivariant networks can be viewed as $n=1$. We use the width of the first layer to denote the network width $c$, since the width of the following layers scales with the first layer proportionally. As feature maps play a major role in memory consumption, $c\times n$ gives a rough idea of the memory cost of a model. 

\begin{table}[]
\resizebox{\columnwidth}{!}{
\begin{tabular}{l|ccccc}
\toprule
Method                                                         & $LSTQ$        & $S_{assoc}$   & $S_{cls}$     & $IOU^{St}$    & $IOU^{Th}$    \\ \midrule
RangeNet++\cite{milioto2019rangenet++}+PP+MOT & 43.8          & 36.3          & 52.8          & 60.5          & 42.2          \\
KPConv\cite{thomas2019kpconv}+PP+MOT          & 46.3          & 37.6          & 57.0          & 64.2          & 54.1          \\
RangeNet++\cite{milioto2019rangenet++}+PP+SFP & 43.4          & 35.7          & 52.8          & 60.5          & 42.2          \\
KPConv\cite{thomas2019kpconv}+PP+SFP          & 46.0          & 37.1          & 57.0          & 64.2          & 54.1          \\
4D-DS-Net\cite{hong2022lidar}                 & 68.0          & 71.3          & \textbf{64.8}          & 64.5          & 65.3          \\
4D-PLS\cite{aygun20214d}                      & 62.7          & 65.1          & 60.5          & 65.4          & 61.3          \\
4D-StOP\cite{kreuzberg20234d}                 & 67.0          & 74.4          & 60.3          & 65.3          & 60.9          \\ \midrule
Eq-4D-PLS \textit{(ours)}                     & 65.0          & 67.7          & 62.3          & 66.4          & 64.6          \\
Eq-4D-StOP \textit{(ours)}                   & \textbf{70.1} & \textbf{77.6} & 63.4 & \textbf{66.4} & \textbf{67.1} \\ \bottomrule
\end{tabular}
}
\caption{SemanticKITTI validation set result. PP: PointPillars \cite{lang2019pointpillars}. MOT: tracking-by-detection by \cite{weng20203d}. SFP: tracking-by-detection with scene flow \cite{mittal2020just}. The best is highlighted in \textbf{bold}. }
\label{tab:val}
\end{table}

\begin{table}[]
\resizebox{\columnwidth}{!}{
\begin{tabular}{l|ccccc}
\toprule
Method            & $LSTQ$ & $S_{assoc}$ & $S_{cls}$ & $IoU^{St}$ & $IoU^{Th}$ \\ \midrule
RangeNet++\cite{milioto2019rangenet++}+PP+MOT & 35.5 & 24.1     & 52.4   & 64.5       & 35.8        \\
KPConv\cite{thomas2019kpconv}+PP+MOT     & 38.0 & 25.9     & 55.9   & 66.9       & 47.7        \\
RangeNet++\cite{milioto2019rangenet++}+PP+SFP & 34.9 & 23.3     & 52.4   & 64.5       & 35.8        \\
KPConv\cite{thomas2019kpconv}+PP+SFP     & 38.5 & 26.6     & 55.9   & 66.9       & 47.7        \\
4D-PLS\cite{aygun20214d} & 56.9 & 56.4     & 57.4   & 66.9       & 51.6        \\
4D-DS-Net\cite{hong2022lidar}         & 62.3 & 65.8     & 58.9   & 65.6       & 49.8        \\
CA-Net\cite{marcuzzi2022contrastive}            & 63.1 & 65.7     & 60.6   & 66.9       & 52.0        \\
4D-StOP\cite{kreuzberg20234d} & 63.9 & 69.5     & 58.8   & 67.7       & 53.8        \\ \midrule
Eq-4D-StOP \textit{(ours)} & \textbf{67.8} & \textbf{72.3}     & \textbf{63.5}   & \textbf{70.4}       & \textbf{61.9}        \\  \bottomrule
\end{tabular}
}
\caption{SemanticKITTI test set result. }
\label{tab:test}
\end{table}

\subsection{Quantitative Results}\label{sec:exp_res}
The evaluation results of our equivariant models on the SemanticKITTI validation set are shown in \cref{tab:val}. 
Compared with 4D-PLS, our equivariant model improves by \textbf{2.3} points on $LSTQ$. Our Eq-4D-StOP model outperforms its non-equivariant baseline by \textbf{3.1} $LSTQ$ points. The Eq-4D-StOP model achieves state-of-the-art performance among published methods. From these experiments, we have the following observations:
\begin{itemize}
    \item The improvements on $IoU^{St}$ are similar for both Eq-4D-StOP and Eq-4D-PLS compared with their non-equivariant baselines.
    \item The improvements on $IoU^{Th}$ and $S_{assoc}$ are larger than on $IoU^{St}$. 
    \item Eq-4D-StOP gains larger improvements on $IoU^{Th}$ and $S_{assoc}$ than Eq-4D-PLS does over their non-equivariant baselines. 
\end{itemize}
The observations indicate the following. First, the introduction of equivariance brings improvements to both models across all metrics consistently. Second, objects (the \textit{things} classes) enjoy more benefits from the equivariant models compared with the background (the \textit{stuff} classes). We hypothesize that this is because objects present more rotational symmetry as compared to background classes. Third, the fact that more significant improvements are observed in Eq-4D-StOP shows the benefit of formulating the equivariant vector field regression problem induced from the offset-based clustering strategy. Improved clustering directly benefits the instance segmentation, i.e., $S_{assoc}$, and it also improves the semantic segmentation of objects ($IoU^{Th}$), since 4D-StOP unifies the semantic class prediction of all points belonging to a single instance by majority voting. 

The evaluation results on the SemanticKITTI test set are shown in \cref{tab:test}. Eq-4D-StOP achieves \textbf{3.9} points improvement in $LSTQ$ over the non-equivariant model and ranks 1st in the leaderboard at the time of submission. 
We only test our best model on the test set, thus the result of Eq-4D-PLS is not available.

We use max-pooling in Eq-4D-PLS and average pooling in Eq-4D-StOP as the invariant pooling layer. The ablation study is in \cref{sec:abl_layer}. The Eq-4D-PLS model is with $c=128, n=6$, and the Eq-4D-StOP model is with $c=128, n=4$. These parameters are selected based on the network scaling analysis in \cref{sec:abl_eff}. 

\begin{table}[]
\centering
\resizebox{0.85\columnwidth}{!}{
\begin{tabular}{c|c|cc}
\toprule
\multirow{2}{*}{\begin{tabular}[c]{@{}c@{}}Target \\ type\end{tabular}} & \multirow{2}{*}{\begin{tabular}[c]{@{}c@{}}Layer \\ type\end{tabular}} & \multicolumn{2}{c}{$LSTQ$}    \\ \cmidrule{3-4} 
                                                                        &                                                                        & Eq-4D-StOP    & Eq-4D-PLS     \\ \midrule
\multirow{4}{*}{Invariant}                                              & Max                                                                    & 68.2          & \textbf{63.7} \\
                                                                        & Average                                                                & \textbf{69.2} & 61.4          \\
                                                                        & Attentive                                                              & 68.5          & 63.1          \\
                                                                        & RCS                                                                    & 68.4          & n/a             \\ \midrule
\multirow{2}{*}{Equivariant}                                            & RCS                                                                    & \textbf{69.2} & n/a             \\
                                                                        & Average                                                                & 61.6          & n/a             \\ \bottomrule
\end{tabular}
}
\caption{Ablation study on the invariant pooling layers and rotational coordinate selection (RCS). All comparisons use $c=128, n=3$. Some options are n/a for Eq-4D-PLS (see \cref{sec:seg_inv}). }
\label{tab:abl}
\end{table}

\subsection{Ablation Study}\label{sec:abl_layer}
\paragraph{Invariant pooling and RCS}
In \cref{tab:abl}, we show the segmentation performance with different invariant pooling layers and the effect of rotational coordinate selection (RCS). In terms of the invariant pooling layers, average pooling is optimal in Eq-4D-StOP, while max pooling is best in Eq-4D-PLS. Eq-4D-PLS favors max pooling, possibly because the point embeddings averaged over all rotational directions may be less discriminative, hindering instance clustering. In contrast, Eq-4D-StOP benefits from average pooling gathering information from all orientations, as the pooled features are used for object property prediction instead of clustering. 
For the equivariant field (offset) prediction, we compare the rotational coordinate selection with average pooling that wrongly treats the offsets as rotation-invariant targets. The performance drastically decreases when RCS is replaced with average pooling, showing that it is of vital importance to respect the equivariant nature of the targets. 

\subsection{Generalization on the nuScenes dataset}
We extend experiments to the nuScenes \cite{fong2022panoptic} dataset. The hyperparameters follow the experiments on SemanticKITTI to show the generalizability of our proposed method. As shown in \cref{tab:nuscenes}, our model (with $c=128, n=4$) largely outperforms the baseline (with $c=512$), further validating the value of our equivariant approach. 

\begin{table}[]
\centering
\resizebox{\linewidth}{!}{
\begin{tabular}{lccccc}
\toprule
Method     & $LSTQ$ & $S_{assoc}$ & $S_{cls}$ & $IOU^{St}$ & $IOU^{Th}$ \\ \midrule
4D-StOP    & 60.5   & 62.5        & 58.6      & 74.4       & 54.9       \\ 
Eq-4D-StOP (\textit{ours}) & \textbf{67.3}   & \textbf{73.7}        & \textbf{61.5}      & \textbf{76.4}       & \textbf{58.7}      \\
\bottomrule
\end{tabular}
}
\caption{4D panoptic segmentation on nuScenes. The models are trained on the training split and evaluated on the validation split. }
\label{tab:nuscenes}
\end{table}

\begin{table}[]
\centering
\resizebox{0.9\columnwidth}{!}{
\begin{tabular}{c|ccccc}
\toprule
Anchor size $n$           & 1    & 2    & 3    & 4    & 6    \\ \midrule
Performance ($LSTQ$)        & 67.1 & 67.3 & 69.3 & 69.8 & 69.0 \\
Inference speed (fps) & 0.73 & 1.06 & 1.14 & 1.27 & 1.39 \\ \bottomrule
\end{tabular}}
\caption{Running speed analysis of Eq-4D-StOP, given constant feature map size $c\times n = 256$. Note that $n=1$ corresponds to the non-equivariant baseline. }
\label{tab:fps}
\end{table}

\begin{figure}
    \centering
    \includegraphics[width=\columnwidth]{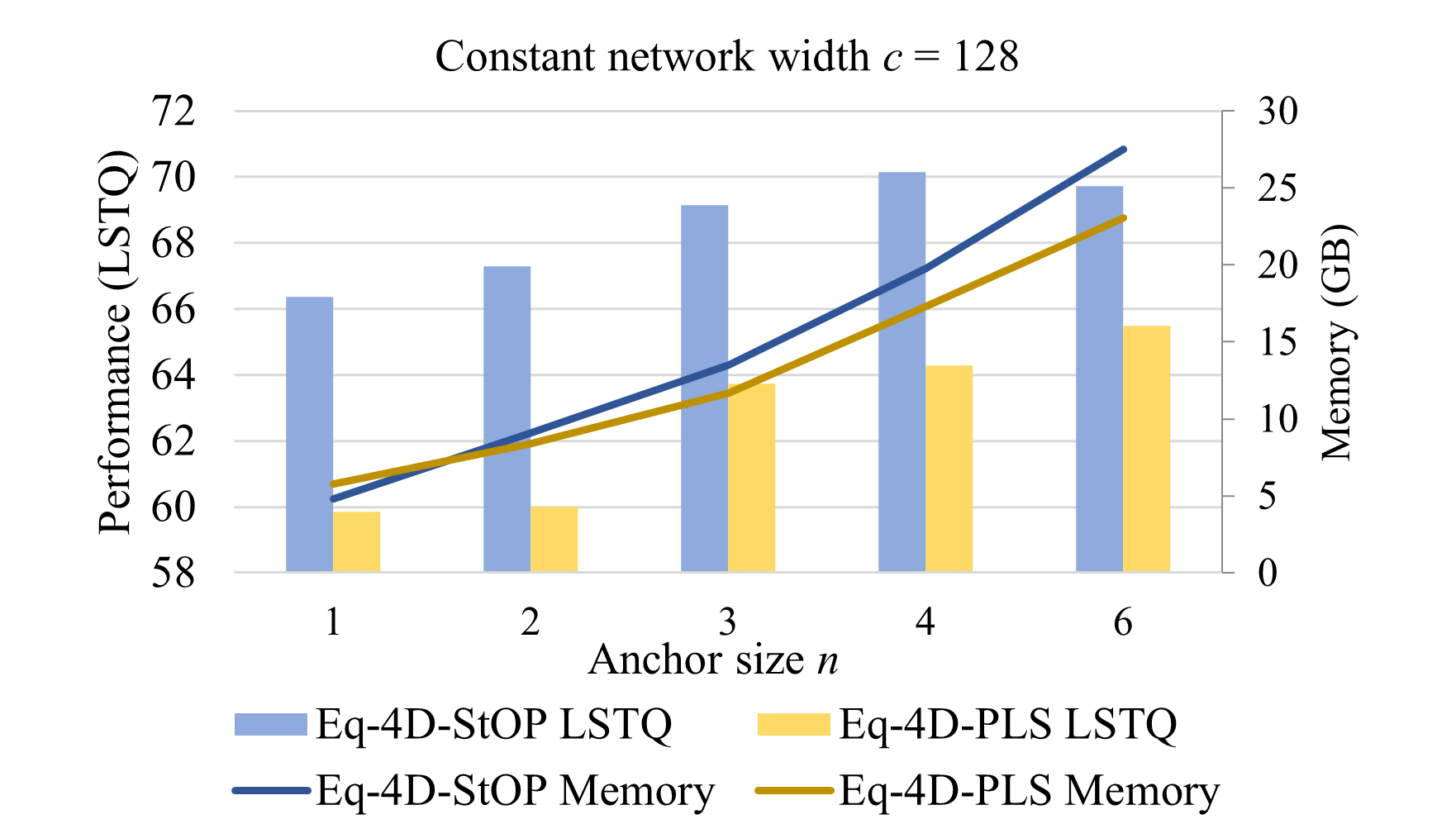}
    \caption{Effect of anchor size $n$ on performance and memory usage in training, given constant network width $c$. Note that $n=1$ corresponds to the non-equivariant baselines. }
    \label{fig:const_width}
\end{figure}

\begin{figure}
    \centering
    \includegraphics[width=\columnwidth]{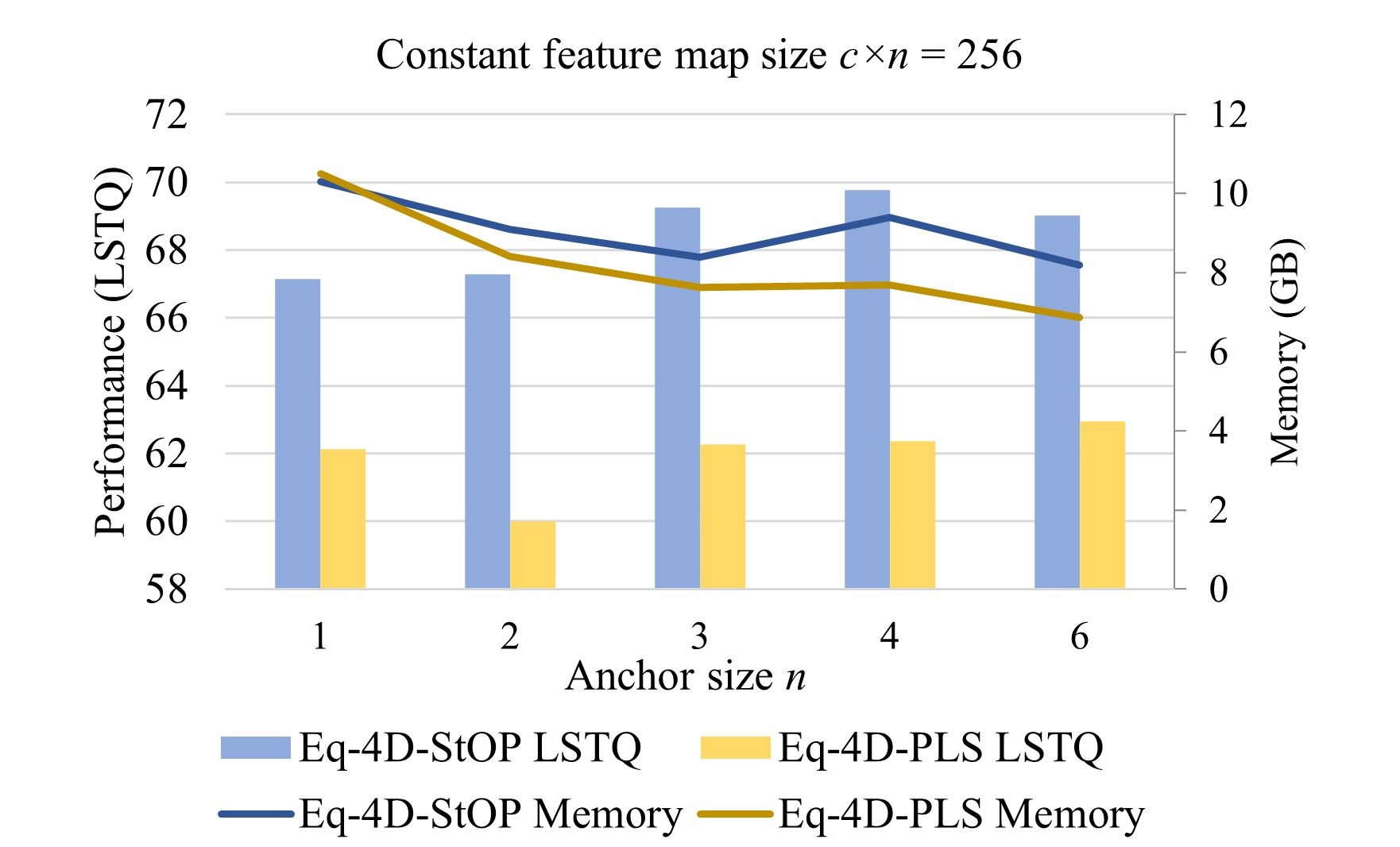}
    \caption{Effect of anchor size $n$ on performance and memory usage in training, given a constant size of feature maps $c\times n$. }
    \label{fig:const_size}
\end{figure}

\begin{figure}
    \centering
    \includegraphics[width=\columnwidth]{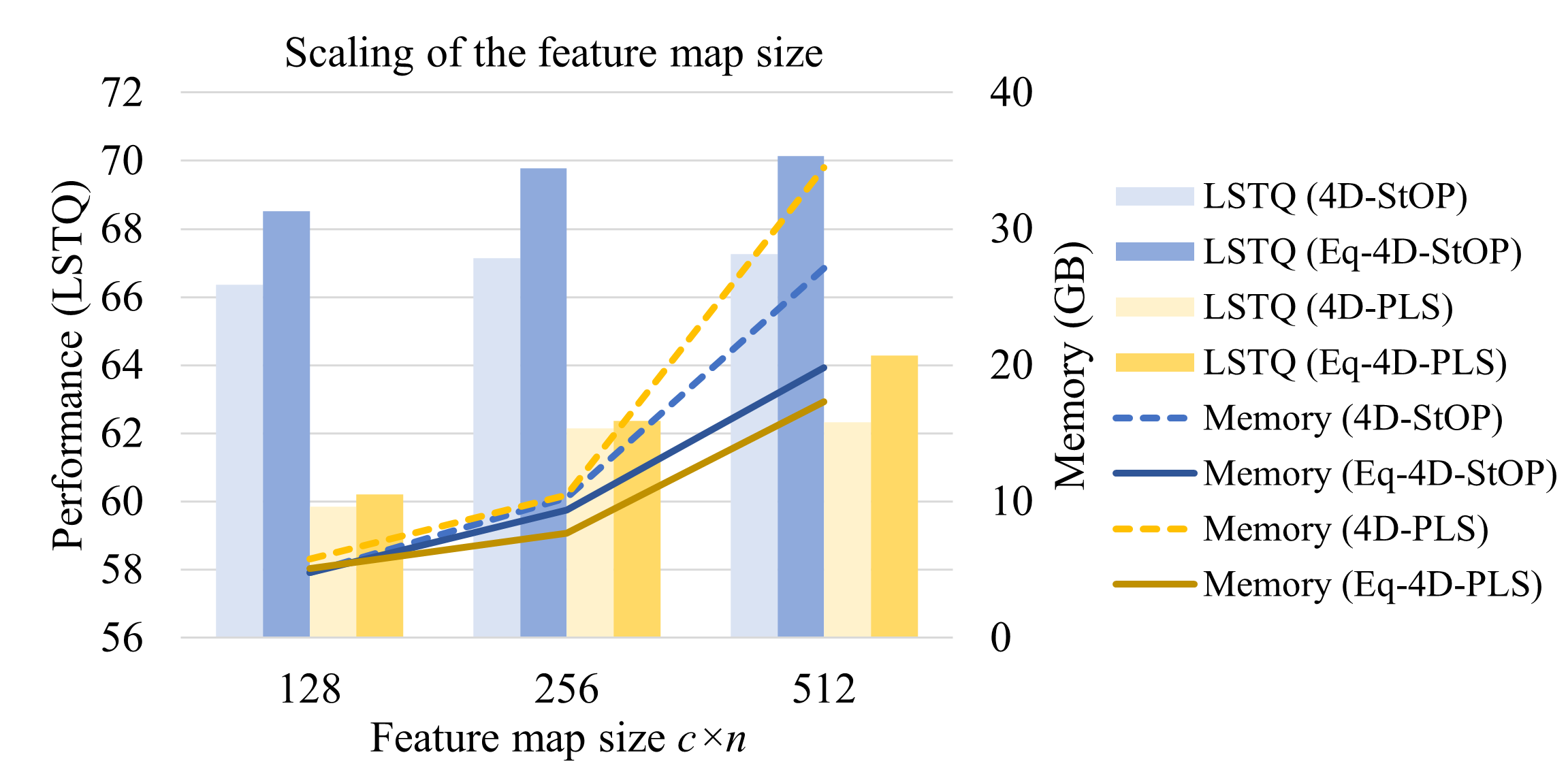}
    \caption{Comparison between equivariant and non-equivariant models at different network sizes. }
    \label{fig:scaling}
\end{figure}

\subsection{Network Scaling and Computational Cost}\label{sec:abl_eff}
We experiment on different network widths $c$ and anchor sizes $n$ to investigate the effect of network sizes on the model performance and efficiency. The $LSTQ$ is evaluated on the SemanticKITTI validation set. 

In \cref{fig:const_width}, with $c=128$, we test various anchor sizes $n$. A larger $n$ more closely approximates continuous $\mathrm{SO}(2)$ equivariance but also proportionally expands feature maps and memory usage. The performance of both Eq-4D-PLS and Eq-4D-StOP enhances with increasing $n$, confirming the efficacy of equivariance in this task.

To rule out the factor of varying sizes of feature maps, we keep $c\times n=256$ with different combinations of $c$ and $n$ (for $n=3$, we use $c=85$ as an approximation). As shown in \cref{fig:const_size}, Eq-4D-StOP significantly outperforms the non-equivariant version ($n=1$) when $n>=3$. The memory consumption even decreases when we increase $n$. There are two reasons. First, the size of the weight matrix in convolution layers gets smaller. Denote the number of kernel points as $k$, for a convolution layer with input and output width $c$, the size of the weight matrix is $knc^2$. Given constant $c\times n$, this number decreases with larger $n$. Second, the feature maps after the invariant pooling layer are smaller (of size $m\times c$). The small bump-up of memory usage at $n=4$ is due to the larger convolution kernel ($k=19$ v.s. $k=15$ as discussed in \cref{sec:encdec}), but it is still lower than the memory usage of the non-equivariant baseline. With constant feature map size $c\times n$, there is a trade-off between the number of rotation anchors $n$ and the feature channels per rotation anchor, which explains the slight performance decline at $n=6$. A similar trend can also be observed in Eq-4D-PLS, but with a smaller performance margin. 

We also report the running time of our models in \cref{tab:fps}. The equivariant models run faster with higher accuracy. 

We further investigate whether the advantage of equivariant models only occurs at a specific network size. In \cref{fig:scaling}, we scale the feature map size of the networks. We use $n=4$ for equivariant models in this comparison. The equivariant models outperform the non-equivariant models at all network sizes. The memory consumption increases faster for the non-equivariant models, because the sizes of convolution kernel weight matrices grow quadratically with the network width $c$, while equivariant models have a smaller $c$ given the same feature map size. 

In summary, we found that the equivariant models have better performance than their non-equivariant counterparts with lower computational costs at different network sizes. 

\section{Conclusion}
In this paper, we use equivariant learning to tackle a complicated large-scale perception problem, the 4D panoptic segmentation of sequential point clouds. While equivariant models were generally perceived as more expensive and complex than conventional non-equivariant models, we show that our method can bring performance improvements and lower computational costs at the same time. We also show that the advantage of equivariant models can be better leveraged if we formulate the learning targets as equivariant vector fields, compared with invariant scalar fields. A limitation of this work is that we did not propose drastically new designs on the overall structure of the 4D panoptic segmentation network under the equivariant setup, but it also allows us to conduct apple-to-apple comparisons regarding equivariance on this task so that our contribution is orthogonal to the improvements in the specific network design. We hope our work could inspire wider incorporation of equivariant networks in practical robotic perception problems. 

\begin{appendix}
\clearpage
\begin{center}
{\bf \Large Appendix}
\end{center}
    
\section{Visualization of Offset Prediction}
\begin{figure}
    \centering
    \resizebox{\columnwidth}{!}{
    \includegraphics{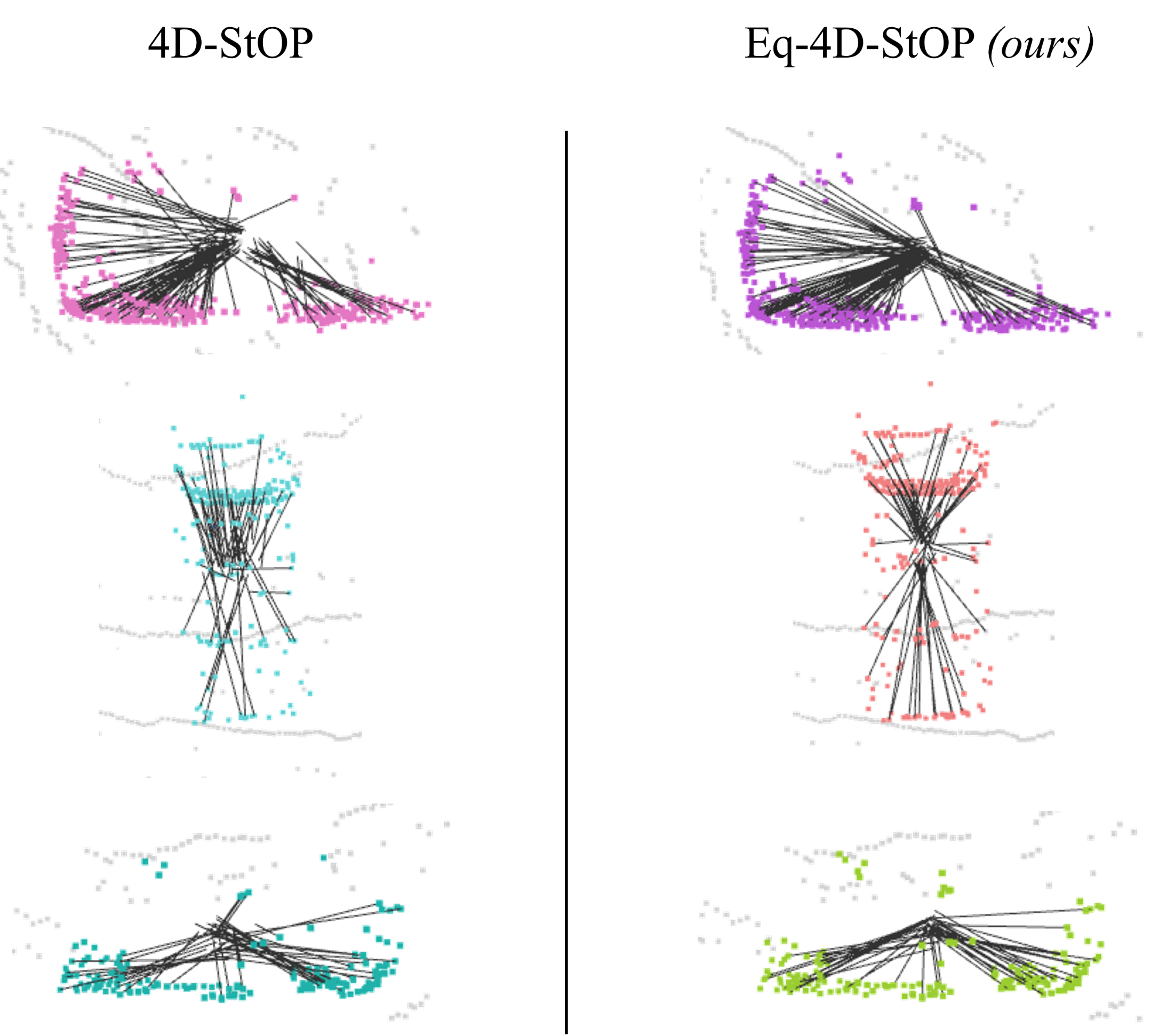}
    }
    \caption{Qualitative comparison of the offset vector prediction (the black line segments) between our method and the baseline. The predictions from our equivariant model are more consistent and the endpoints are more concentrated near the instance centers. }
    \label{fig:offset}
\end{figure}

The offset prediction as an equivariant vector field is a main factor in the significant improvement achieved by our proposed Eq-4D-StOP model. In \cref{fig:offset}, we visualize the offset predictions to show this improvement intuitively. We can see that the offset vectors predicted by our equivariant model have more consistent orientations and the end points are closer to the instance center, thus benefiting the object clustering and segmentation. 

\section{3D Panoptic Segmentation Performance}
While the main focus of this paper is 4D panoptic segmentation, the network structure is also compatible with the 3D panoptic segmentation task by skipping the point cloud aggregation step and only taking a single scan of point cloud as input. In the 3D panoptic segmentation task, we keep the model and training configurations the same as in Sec. 5.2, except for inputting a single frame of point cloud during training and inference. 
In \cref{tab:3d}, we show the performance of our model compared with the baseline. The metrics follow the 2D \cite{kirillov2019panoptic, porzi2019seamless} and 3D \cite{li2022panoptic, gasperini2021panoster} panoptic segmentation literature. $PQ$, the panoptic quality, measures the overall accuracy of panoptic segmentation. $PQ=SQ\times RQ$, where $RQ$, the recognition quality, measures the ratio of successful instance segmentation with $IoU>0.5$, and $SQ$ measures the segmentation quality by the average $IoU$ across the successfully segmented instances. The superscripts $Th$ and $St$ refer to the \textit{things} classes and \textit{stuff} classes, as in the 4D metrics. The semantic segmentation accuracy is measured by $mIoU$. 

From \cref{tab:3d}, we can see that the performance of our Eq-4D-StOP model improves over the non-equivariant baseline in all metrics, which shows that the equivariance property also benefits the 3D panoptic segmentation task. Especially, $PQ^{Th}$ is increased by 4.0 points, showing that the instance segmentation of objects is majorly improved, consistent with our observations in Sec. 5.2 in the 4D segmentation. 

\begin{table*}[]
    \centering
    \begin{tabular}{l|cccc|ccc|ccc|c}
    \toprule
    Method            & $PQ$ & $PQ^{\dagger}$ & $SQ$ & $RQ$ & $PQ^{Th}$ & $SQ^{Th}$ & $RQ^{Th}$ & $PQ^{St}$ & $SQ^{St}$ & $RQ^{St}$ & $mIoU$ \\ \midrule
    4D-StOP \cite{kreuzberg20234d} & 58.5 & 64.0           & 80.3 & 68.2 & 62.1      & 91.0      & 67.8      & 56.0      & 72.5      & 68.6      & 64.6   \\
    Eq-4D-StOP \textit{(ours)} & 61.2 & 66.2           & 83.6 & 70.8 & 66.1      & 91.3      & 71.9      & 57.5      & 78.0      & 70.0      & 68.0   \\ \bottomrule
    \end{tabular}
    \caption{3D Panoptic segmentation results on SemanticKITTI validation set. }
    \label{tab:3d}
\end{table*}

\section{Ablation: Rotation Classification for Offset Prediction without Equivariant Features}
Besides the benefits brought by the equivariance, there could be another hypothesis for the performance improvement in Eq-4D-StOP: With the rotation classification, it could be easier to regress the offset vector. It can be explained as follows. As discussed in Sec. 4.3.2, for a point $x\in \mathbb{R}^3$ with arbitrary target offset vector $v \in \mathbb{R}^3$, its corresponding orientation is $\theta(v) = \mathrm{atan2}(v_Y, v_X)$. Here we slightly abuse the notation to use $\theta$ to represent both the angle and the corresponding rotation matrix. It should not cause ambiguity since all rotations are in $\mathrm{SO}(2)$ in this discussion. The ground truth rotation anchor for vector $v$ is $\theta_{i(v)} \in \mathrm{SO}(2)'$, where $i(v) = \arg \min_{i} \norm{\theta_{i} -\theta(v)}$. Following Eq.~(4) and (6), the learning process is to fit $f(x, \theta_{i(v)})$, the prediction at the $i(v)$'th rotation anchor, to $\theta_{i(v)}^{-1} v$. Intuitively speaking, it means that the offset is always regressed in the local reference frame (rotation anchor) closest to the orientation defined by the offset itself. The variation of $v$ in its closest local reference frame is much smaller than $v$ in the global frame. Specifically, $\theta(\theta_{i(v)}^{-1}v) = \theta_{i(v)}^{-1}\theta(v) \in [-\frac{\pi}{n}, \frac{\pi}{n} )$, for $SO(2)'\cong C_n$ with $n$ discretized rotation anchors. In comparison, $\theta(v)\in [-\pi, \pi)$, which implies that the regression of $\theta_{i(v)}^{-1}v$ could be easier. 

We test out this hypothesis by experimenting with a network that uses the non-equivariant KPConv \cite{thomas2019kpconv} backbone and predicts the offset with rotation classification. The ground truth rotation anchors are defined in the same way as the equivariant models, and the target offset to be regressed is also $\theta_{i(v)}^{-1}v$ as discussed above. We call this model 4D-StOP with rotation head (R-head), as in the last column of \cref{tab:rot_head}, which compares the performance with the baseline and our equivariant model. The comparison uses a consistent feature map size and rotation anchor size. The experimental results show that 4D-StOP w/ R-head does not outperform the baseline, indicating that the performance improvement is brought by the equivariant property of the network instead of the smaller variations in the regression targets. 

\begin{table}[]
\centering
\resizebox{\linewidth}{!}{
\begin{tabular}{lccc}
\toprule
Method & \begin{tabular}[c]{@{}c@{}}4D-StOP \\ ($c=256$)\end{tabular} & \begin{tabular}[c]{@{}c@{}}Eq-4D-StOP \\ ($c=64, n=4$)\end{tabular} & \begin{tabular}[c]{@{}c@{}}4D-StOP w/ \textit{R-head} \\ ($c=256, n=4$)\end{tabular} \\ \midrule
$LSTQ$ & 67.1                                                         & \textbf{69.8}                                                                & 66.8         \\
\bottomrule
\end{tabular}
}
\vspace{-2pt}
\caption{Experiment of standard KPConv and proposed prediction head for equivariant field prediction (\textit{R-head}) on SemanticKITTI.  }
\label{tab:rot_head}
\vspace{-5pt}
\end{table}

\section{Quotient Representation in SO(2) Causes Information Loss}\label{sec:quo}
In Sec. 4.2, we introduce that we use the regular representation instead of the quotient representation in our $\mathrm{SO}(2)$ equivariant 4D panoptic segmentation network, because quotient representations cause information loss for abelian groups like $\mathrm{SO}(2)$. Here is a more detailed explanation. 

First, we explain what it means to have a quotient representation that does \textit{not} cause information loss, as is the case in E2PN \cite{zhu20222}. E2PN is a $\mathrm{SO}(3)$-equivariant network with feature maps in the space $\mathcal{F}=\{f: \mathbb{R}^3 \times S^2 \rightarrow V \}$, where $S^2=\mathrm{SO}(3) / \mathrm{SO}(2)$ is the 2D sphere in 3D space, and also the quotient space of $\mathrm{SO}(3)$ with respect to subgroup $\mathrm{SO}(2)$. As a $\mathrm{SO}(3)$-equivariant network, its feature maps are not defined on $\mathrm{SO}(3)$ but only $S^2$, which is why it is said to use a \textit{quotient representation} to reduce the feature map size and thus the computational cost. The reason that this quotient representation does not cause information loss is that the group action of $\mathrm{SO}(3)$ on $S^2$ (i.e., the 3D rotation of a sphere) is \textit{faithful}, which is to say the only rotation in $\mathrm{SO}(3)$ that keeps all points on a sphere unchanged is the identity rotation. It implies that any $\mathrm{SO}(3)$ rotation can be detected from its action on the $S^2$ feature maps, therefore not losing any information in $\mathrm{SO}(3)$. 

Put more formally, we denote the group as $G$, the subgroup as $H$, the quotient space as $G/H$.
The group actions of $G$ on $G/H$ is a group homomorphism $\phi: G \rightarrow Aut(G/H)$. If the group action is faithful, then the kernel of the homomorphism is $ker(\phi) = \{e\}$, only containing the identity element. By the first isomorphism throrem, $G/ker(\phi) = G \cong Im(\phi)$. That is to say, $\phi$ is injective. Therefore, there exists an inverse map $\phi^{-1}: Aut(G/H) \supset Im(\phi) \rightarrow G, \phi(g) \mapsto g$. We can determine the group element $g\in G$ from the automorphism in the quotient space $G/H$, thus we say the information of $G$ is fully preserved in $G/H$. 

However, for $\mathrm{SO}(2)$, which is an abelian group, its action on its quotient space is \textit{not} faithful. To see this, we still use the $G$ to denote $\mathrm{SO}(2)$ and $H$ to denote a subgroup of $G$. An element in the quotient space $G/H$ can be denoted as $gH$ for some $g\in G$. The group action of $G$ on $G/H$ is $g' \mapsto (gH \mapsto g'gH, \forall g\in G)$. Now if we take $g'=h\in H$, then with the abelian property of $G$, we have $g'gH = hgH = ghH = gH$, meaning that the action of $g'\neq e$ on $G/H$ keeps all elements in $G/H$ unchanged. Therefore, the group action of $\mathrm{SO}(2)$ on its quotient space is not faithful, and $ker(\phi) = H$. 

By the first isomorphism throrem, $G/ker(\phi) \cong Im(\phi) \subset G/H$. From $G/H$, we can only recover elements in $G/ker(\phi)=G/H$ instead of $G$, therefore the information inside an $H$-coset is lost. 

Here we provide a concrete example in the discretized case. Consider $\mathrm{SO}(2)$ discretized as $C_6$, i.e., the set composed of 60-degree rotations. If we take a subgroup $C_2$ (i.e., 180-degree rotations), then the quotient space is $C_6 / C_2 = C_3$. From the quotient features $C_3$, we will lost discrimination among the $C_2$-coset. In other words , any rotation angle $\theta$ and $\theta+180^\circ$ corrrespond to the same quotient feature maps in $C_3$. 

Therefore, we use the regular representation instead of the quotient representation. In other words, to enable $C_n$-equivariance, we use a feature map defined on $C_n$ as well. 

\section{Nearest-Neighbor Upsampling and 1-by-1 Convolution Are Equivariant}\label{sec:proof}
\paragraph{Nearest-neighbor upsampling layer}
For the nearest-neighbor upsampling layer, denote a coarse-level feature map as $f_{coarse}\in \mathcal{F}$ and a fine-level feature map as $f_{fine}\in \mathcal{F}$. The nearest neighbor upsampling layer gives
\begin{equation}
    f_{fine}(x, R) = f_{coarse}(x_{nn}, R)
\end{equation}
where $x\in X_{fine}\subset\mathbb{R}^3$, in which $X_{fine}$ is the fine point cloud. $x_{nn}\in X_{coarse}\subset\mathbb{R}^3$, where $X_{coarse}$ the coarse point cloud, and $x_{nn}$ is the nearest neighbor of $x$ in the coarse point cloud. Since distance is preserved under rotations, the nearest neighbor for $Rx$ in the rotated coarse point cloud $RX_{coarse}$ is $Rx_{nn}$. If $f_{coarse}$ is an equivariant feature map, i.e., satisfies Eq. (4), then we have 
\begin{multline}
    [Rf_{fine}](Rx, R') = [Rf_{coarse}](Rx_{nn}, R') \\
    = f_{coarse}(x_{nn}, R^{-1}R') = f_{fine}(x, R^{-1}R'), 
\end{multline}
which means $[f_{fine}]$ also satisfies Eq. (4), thus is equivariant. 

\paragraph{1-by-1 convolution layer}
A 1-by-1 convolution is a map $W: V\rightarrow V$ which operates on $f(x, R)$ for each $x$ and $R$ individually. Denote an existing equivariant feature map $f_1\in \mathcal{F}$ and the feature map after the 1-by-1 feature map $f_2\in \mathcal{F}$, i.e., 
\begin{equation}
    f_2(x, R) = W(f_1(x, R))
\end{equation}
Then we have
\begin{multline}
    [Rf_2](Rx, R') = W([Rf_1](Rx, R')) \\
    = W(f_1(x, R^{-1}R')) = f_2(x, R^{-1}R')
\end{multline}
showing that $f_2$ satisfies Eq. (4), therefore is equivariant. 
\end{appendix}

{\small
\bibliographystyle{ieee_fullname}
\bibliography{egbib}
}

\end{document}